\documentclass[twocolumn]{jingdong}
\usepackage{amsmath,amssymb}
\usepackage{tabularx}
\usepackage{enumitem}
\usepackage{float}

\hypersetup{
  pdftitle={ROSWORLD: When Digital Task Competence Meets Physical Execution},
  pdfauthor={Caoyuan Ma et al.}
}

\fancypagestyle{firststyle}{
  \fancyhf{}
  \fancyhead[L]{\vspace*{2mm}\includegraphics[width=130pt]{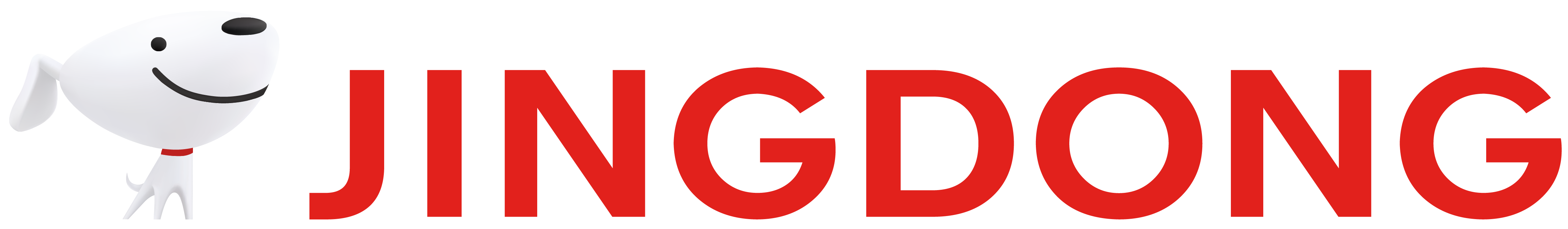}}
  \fancyfoot[C]{\thepage}

}

\newcommand{\benchmark}{\textsc{ROSWorld}}

\newcolumntype{Y}{>{\raggedright\arraybackslash}X}
\setlist{nosep,leftmargin=*}
\title{ROSWORLD: When Digital Task Competence Meets Physical Execution}

\author[1,2,*]{Caoyuan Ma}
\author[1,*]{Wendi Liu}
\author[3]{Zhimin Zhang}
\author[1]{Muyao Niu}
\author[1]{Yujie Lu}
\author[1]{Zengxi Zhang}
\author[4,2]{Ji Zhao}
\author[5,4,2]{Shuai Dong}
\author[3,2]{Wenpu Liu}
\author[3,2]{Ziyue Wang}
\author[2]{Wenzheng Chang}
\author[6,2]{Taiqiang Wu}
\author[3,2]{Weichu Xie}
\author[2]{Yuqi Xu}
\author[7,2]{Haiyuan Wan}
\author[2]{Xuehui Wang}
\author[2]{Jiaqi Wang}
\author[2,\ddagger]{Wenqi Shao}
\author[1,\dagger]{Yinqiang Zheng}

\affiliation[1]{UTokyo}
\affiliation[2]{JD.COM}
\affiliation[3]{PKU}
\affiliation[4]{SII}
\affiliation[5]{FDU}
\affiliation[6]{HKU}
\affiliation[7]{THU}

\contribution[*]{Equal contribution}
\contribution[\ddagger]{Project leader}
\contribution[\dagger]{Corresponding author}

\abstract{
Computer-use agents now solve most desktop tasks through direct software input, but a robot using the same computer must turn GUI knowledge into physical contact with a keyboard and mouse. We introduce \benchmark{}, a benchmark that couples a physically simulated robot workstation to a live OSWorld desktop, so that a robot and a desktop agent pursue the same goals under the same official evaluators. The evaluated model receives observations only from the workstation camera views and operates the peripherals through either a single-arm rigid probe or dual-arm dexterous hands. Across five frontier models and 40 tasks in ten applications, models that complete 31--39 tasks on the desktop complete at most 23 with the probe and 28 with the hands. Paired trajectory analysis shows that models adapt their interaction strategies toward keyboard input, yet physical execution frequently fails through missed presses, incorrect targets, and unintended contacts. \benchmark{} makes the gap between knowing what to do on a computer and physically doing it measurable.
}

\begin{document}
{\let\FloatBarrier\relax\maketitle}
\thispagestyle{firststyle}

\section{Introduction}
\begin{table}[t]
\centering\small
\caption{ROSWorld against related benchmarks. We couple the digital computer-use tasks with embodiment evaluation. \emph{RealVM}: tasks run inside a real VM with live desktop or web applications; \emph{Physical contact}: a simulated or real robot physically interacts with the environment; \emph{Camera view}: the screen is observed only through cameras; \emph{Embodiments}: more than one robot embodiment. \checkmark: yes; $\circ$: partly; --: no.}
\label{tab:related}
\setlength{\tabcolsep}{4pt}
\resizebox{\columnwidth}{!}{%
\begin{tabular}{lcccc}\toprule
Benchmark & RealVM & Physical contact & Camera view & Embodiments \\\midrule
OSWorld~\citep{xie2024osworld} & \checkmark & -- & -- & -- \\
WebArena~\citep{zhou2024webarena} & \checkmark & -- & -- & -- \\
OSWorld-G~\citep{xie2025osworldg} & \checkmark & -- & -- & -- \\
\midrule
AI2-THOR~\citep{kolve2017ai2thor} & -- & $\circ$ & \checkmark & -- \\
Habitat~\citep{savva2019habitat} & -- & $\circ$ & \checkmark & -- \\
ALFRED~\citep{shridhar2020alfred} & -- & $\circ$ & \checkmark & -- \\
BEHAVIOR-1K~\citep{li2024behavior} & -- & \checkmark & \checkmark & -- \\
RLBench~\citep{james2020rlbench} & -- & \checkmark & \checkmark & -- \\
ManiSkill2~\citep{gu2023maniskill2} & -- & \checkmark & \checkmark & -- \\
LIBERO~\citep{liu2023libero} & -- & \checkmark & \checkmark & -- \\
RoboDojo~\citep{chen2026robodojo} & -- & \checkmark & \checkmark & \checkmark \\
EmbodiedBench~\citep{yang2025embodiedbench} & -- & \checkmark & \checkmark & $\circ$ \\
RobotWorld~\citep{yang2026robotworld} & -- & \checkmark & \checkmark & \checkmark \\
Isaac Lab typing~\citep{isaaclab2026typing} & -- & \checkmark & -- & -- \\\midrule
\textbf{ROSWorld (ours)} & \checkmark & \checkmark & \checkmark & \checkmark \\
\bottomrule\end{tabular}%
}
\end{table}

Computer-use agents \citep{shi2017wob,humphreys2022control,shaw2023pix2act} can edit documents, manage files, and configure applications through software-issued keyboard and mouse input. These familiar tasks also offer a rich setting for evaluating models that control robots, where a small set of physical peripherals connects the robot to diverse application workflows, each with a concrete goal. When a model uses a robot to operate the computer, every intended input must be realized through motion and contact. Opening a menu requires locating and actuating the appropriate control; editing a document requires coordinating many such interactions while tracking changes in the application. A missed press or an unintended input can change what must happen next, requiring the model to recognize the discrepancy and recover. Physical computer use therefore brings visual understanding, task planning, precise manipulation, and feedback-based recovery together in sustained, goal-directed behavior. It provides a concrete setting for asking \textbf{how well a model can turn its competence at desktop tasks into effective physical action.}

Many existing embodiment benchmarks \citep{james2020rlbench, gu2023maniskill2, liu2023libero, chen2026robodojo} focus on assessing physical skills with bounded manipulation goals, such as grasping, placing, or actuating an object, while desktop benchmarks \citep{xie2024osworld,bonatti2024windows} provide a rich set of computer-use scenarios with clearly defined, task-oriented purposes. To answer the question above, we propose \benchmark{}, connecting these evaluation surfaces by preserving desktop goals and official outcome evaluators. Models need to complete various desktop applications selected from OSWorld \citep{xie2024osworld} through robot controlling and contact with physical peripherals like keyboard and mouse within a simulated environment. Compared with an isolated manipulation goal, completing a computer task requires the model to maintain reliable physical control over a sequence of state-dependent operations. Closely spaced keys and small actuation travel demand precise contact; modifiers and mouse drags require coordinated holding and release; and the meaning of an input depends on the current window, focus, and editing state. The model must therefore connect the geometry and timing of its bodily actions to their changing effects in the digital world, recognize ineffective or unintended inputs, and recover while preserving progress toward the task goal.

\benchmark{} evaluates end-to-end physical computer use under a fixed observation and action interface. Perception, action selection, robot control, physical contact, and feedback-based recovery are parts of a single embodied task rather than separately evaluated capabilities. This preserves the concrete challenges of physical computer use: reading the display from camera views, translating GUI intent into accurate 3D contact, and verifying state changes to recover over long horizons, while shared OSWorld tasks and evaluators allow direct comparison with software-based computer use.

We couple an Isaac Sim workstation to a live OSWorld desktop. The model reads camera views and robot status, then produces bounded motion sequences. Contact with the keyboard and mouse generates desktop input, and the updated desktop is rendered on the workstation monitor for the next observation.

We evaluate five models on 40 shared tasks across ten application domains with two embodiment configurations. Aligning each model's desktop and physical trajectories shows that the task route is preserved, whereas the input modality changes. Models tend to convert mouse operations into keyboard operations in both embodiment configuration, and with the probe they look for single-key substitutes for key chords. Most physical failures arise from presses that do not reach the key, presses on the wrong target, and unintended contacts made by other parts of the robot. These unintended contacts reveal that some models fail to account for the robot's full body during interaction. At the task level, most failed runs exhaust their step budget, some give up when they judge the task infeasible, and only a few wrongly declare the task complete without passing the evaluator.

Our contributions are:
\begin{itemize}
\item A benchmark that couples simulation with a real system: input signals generated by robot contact with the keyboard and mouse in simulation drive a live VM, yielding long-horizon, fine-grained, and verifiable tasks.
\item A paired evaluation of five models on 40 OSWorld tasks under shared goals and official evaluators, comparing direct desktop input with two robot embodiments to quantify the gap between desktop competence and physical completion.
\item A trajectory analysis that links model plans, robot contact, device events, and application outcomes to characterize interaction strategies and identify failures at both the operation and task levels.
\end{itemize}

\section{Related Work}
Table~\ref{tab:related} positions \benchmark{} among related benchmarks. It is the first benchmark which bridges computer-use tasks and embodiment evaluation. The model produces desktop input through physical contact with a keyboard and mouse via robotic control, observes the screen only through camera views, and is scored by the same application-state evaluators as the paired digital run.

\paragraph{GUI agents and computer-use benchmarks.}
GUI agents combine visual grounding with sequential interaction.
Early work learns screenshot-based keyboard and mouse control
\citep{shi2017wob,humphreys2022control,shaw2023pix2act}, while ScreenAgent, Cradle, UFO,
Agent S, and AssistGUI develop multi-step computer-control frameworks
\citep{niu2024screenagent,tan2024cradle,zhang2024ufo,agashe2024agents,gao2023assistgui}.
SeeAct and WebVoyager address web interaction \citep{zheng2024seeact,he2024webvoyager}.
SeeClick and UGround localize interface elements
\citep{cheng2024seeclick,gou2025uground}; Ferret-UI, ScreenAI, and OmniParser study screen understanding and parsing
\citep{you2024ferretui,baechler2024screenai,lu2024omniparser}; ScreenSpot-Pro evaluates high-resolution grounding \citep{li2025screenspotpro}.
CogAgent, ShowUI, Aguvis, and UI-TARS integrate visual understanding and action
\citep{hong2023cogagent,lin2024showui,xu2025aguvis,qin2025ui}.
OSWorld and Windows Agent Arena evaluate desktop tasks
\citep{xie2024osworld,bonatti2024windows}; OmniACT evaluates executable task
scripts \citep{kapoor2024omniact}. Mind2Web provides web demonstrations
\citep{deng2023mind2web}; WebArena, VisualWebArena, and WorkArena evaluate
interactive web workflows \citep{zhou2024webarena,koh2024visualwebarena,drouin2024workarena};
AndroidWorld checks mobile task outcomes \citep{rawles2024androidworld}.
OSWorld extensions examine complementary aspects of computer use:
OSWorld-Verified improves task quality and evaluator robustness
\citep{xlang2025osworldverified}; OSWorld-G isolates GUI grounding, including
layout understanding and precise manipulation \citep{xie2025osworldg};
OSWorld-Human compares agent trajectories with human references to assess
interaction efficiency \citep{abhyankar2025osworldhuman}; and OSWorld-MCP
jointly evaluates GUI interaction and MCP tool invocation
\citep{jia2026osworldmcp}.
We select tasks from OSWorld-1 on which mainstream models already demonstrate strong desktop performance (Table~\ref{tab:results}), providing a capable digital reference for studying physical execution. \benchmark{} retains these goals and official evaluators while replacing direct software input with robot sensing and contact, enabling a paired comparison of the same model on the same task.

\paragraph{Embodied evaluation.}
Embodied evaluation spans simulation platforms, instruction-following tasks,
and robot manipulation. AI2-THOR provides interactive indoor scenes, while
Habitat supports scalable visual navigation and embodied-agent evaluation
\citep{kolve2017ai2thor,savva2019habitat}.
ALFRED evaluates compositional household tasks from language instructions
and egocentric observations \citep{shridhar2020alfred}, and BEHAVIOR-1K
extends evaluation to a broad range of everyday activities
\citep{li2024behavior}.
For robot manipulation, RLBench supplies a multi-task learning environment
\citep{james2020rlbench}; Meta-World studies multi-task and meta-learning
across manipulation goals \citep{yu2019metaworld}; ManiSkill2 emphasizes
skills that generalize across object geometry and manipulation settings
\citep{gu2023maniskill2}; LIBERO studies knowledge transfer in lifelong robot manipulation \citep{liu2023libero}.
More recently, EmbodiedBench studies multimodal agents across semantic and
low-level embodied tasks \citep{yang2025embodiedbench}, while RobotWorld
examines general-purpose agents across diverse robot embodiments and physical
goals \citep{yang2026robotworld}. RoboDojo evaluates generalist manipulation policies in simulation and the real world, with simulation tasks targeting generalization,
memory, precision, long-horizon execution, and open-vocabulary instruction
following \citep{chen2026robodojo}.
These works characterize embodied generalization. Our focus is a matched
comparison of the same model and desktop goal under software input and robot
contact, using a shared official outcome evaluator.

Robot-video benchmarks also study fine-grained event understanding and trajectory evaluation. WGO-Bench evaluates temporal segmentation and semantic labeling of manipulation events \citep{macrodata2026wgobench}, while RoboRewardBench, Robometer, and Flora study outcome assessment, progress estimation, and reward annotation \citep{lee2026roboreward,liang2026robometer,leng2026flora}. RubricBench studies rubric-guided evaluation of text responses \citep{zhou2026rubricbench}. These works inform trajectory analysis and evaluator design; \benchmark{} evaluates agents acting in a closed loop, with success determined by live application state.

\paragraph{Vision-language-action models.}
Vision-language-action (VLA) models such as RT-2, OpenVLA, and $\pi_0$
learn robot actions from visual observations, language instructions, and
manipulation demonstrations \citep{brohan2023rt2,kim2024openvla,black2024pi0}.
SayCan grounds language planning in skill affordances \citep{ahn2022saycan},
while PaLM-E incorporates continuous sensory inputs into language models
\citep{driess2023palme}. Inner Monologue uses feedback for embodied planning
\citep{huang2022innermonologue}, and VoxPoser grounds manipulation in spatial value maps
\citep{huang2023voxposer}. \benchmark{} evaluates general multimodal models
with desktop competence, requiring them to execute application workflows
through physical keyboard and mouse interaction.

\paragraph{Robot typing and paired desktop goals.}
Robot typing includes tactile reinforcement learning on braille keycaps
\citep{church2020typing} and hierarchical reinforcement learning for dual-arm
keyboard input \citep{baltes2024typing}. The Isaac Lab SO-101 project trains
letter presses using a fixed key-position map and provides hardware deployment
\citep{isaaclab2026typing}. \benchmark{} connects camera-observed typing,
mouse manipulation, and application workflows to live application-state scores.
Whereas RobotWorld compares tasks and embodiments \citep{yang2026robotworld},
\benchmark{} pairs digital and physical execution of the same desktop goal
by the same model under a shared evaluator.

\section{ROSWORLD}
\label{sec:environment}
\subsection{Shared digital goals, physical interaction}
\begin{figure*}[t]
\centering
\includegraphics[width=\textwidth]{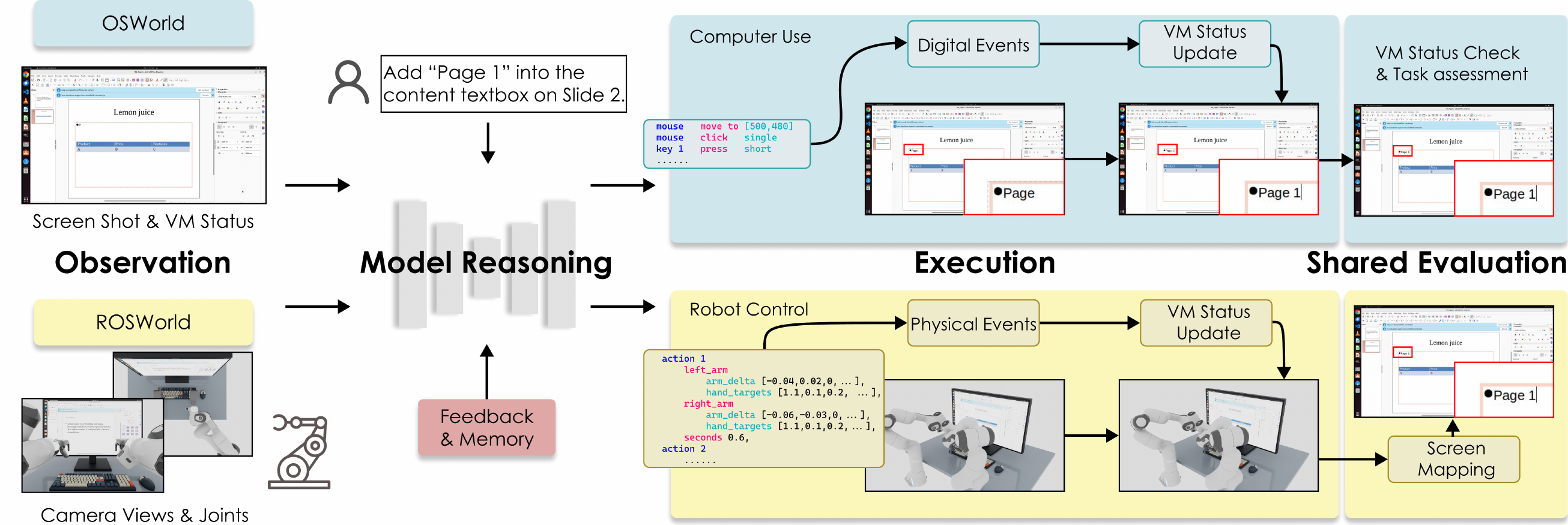}
\caption{Overview of ROSWORLD. The live OSWorld desktop is rendered on the simulated monitor. The model observes camera views and controls the robot to interact with the keyboard and mouse. Contact-generated input events are delivered to the VM, closing the physical--digital loop. Task success is scored by OSWorld's official application-state evaluators.}
\label{fig:loop}
\end{figure*}

A workstation contains a monitor, keyboard, mouse, and Franka robot.
Isaac Sim and PhysX simulate robot and peripheral motion at 480\,Hz \citep{nvidiaIsaacSim,nvidiaPhysX}.
The monitor displays a live Ubuntu VM initialized by OSWorld. Physical key
travel and mouse motion generate desktop input; the resulting desktop image
updates the monitor texture. Figure~\ref{fig:loop} illustrates this feedback loop.

We evaluate two embodiments. A single Franka with a rigid probe has a compact
contact tip for arm positioning, key actuation, and mouse manipulation, but it
cannot hold one key while pressing another. The dual-arm embodiment uses two Franka arms, each equipped with a 6-DoF Inspire Hand, enabling key chords and button-held gestures. Both embodiments use the same desktop
tasks and physical peripheral assets.

\subsection{Observations and actions}
\begin{figure}[htbp]
\centering
\begin{minipage}{0.51\linewidth}
\includegraphics[width=\linewidth]{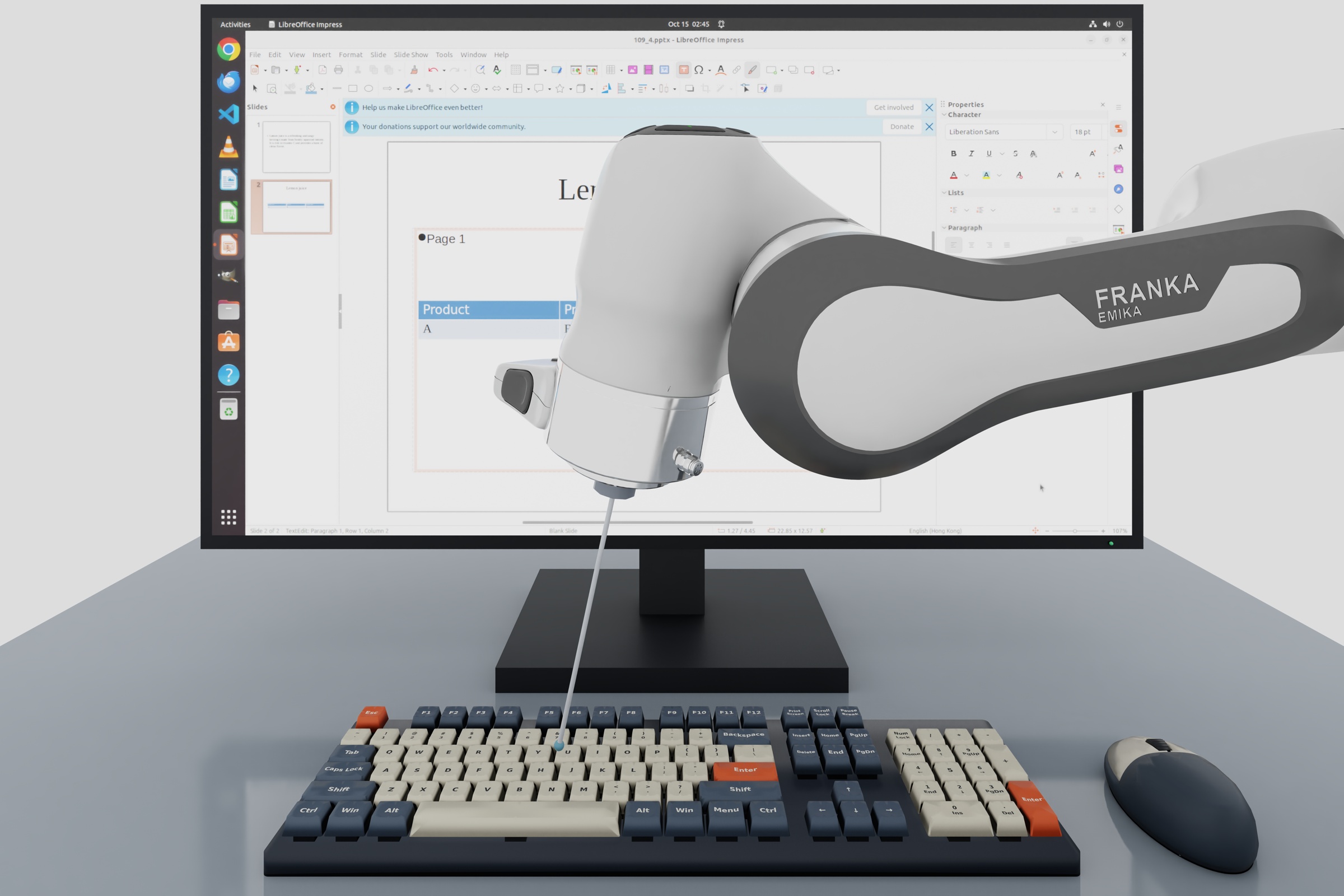}\\
\centering\small (a) Front view
\end{minipage}\hfill
\begin{minipage}{0.46\linewidth}
\includegraphics[width=\linewidth]{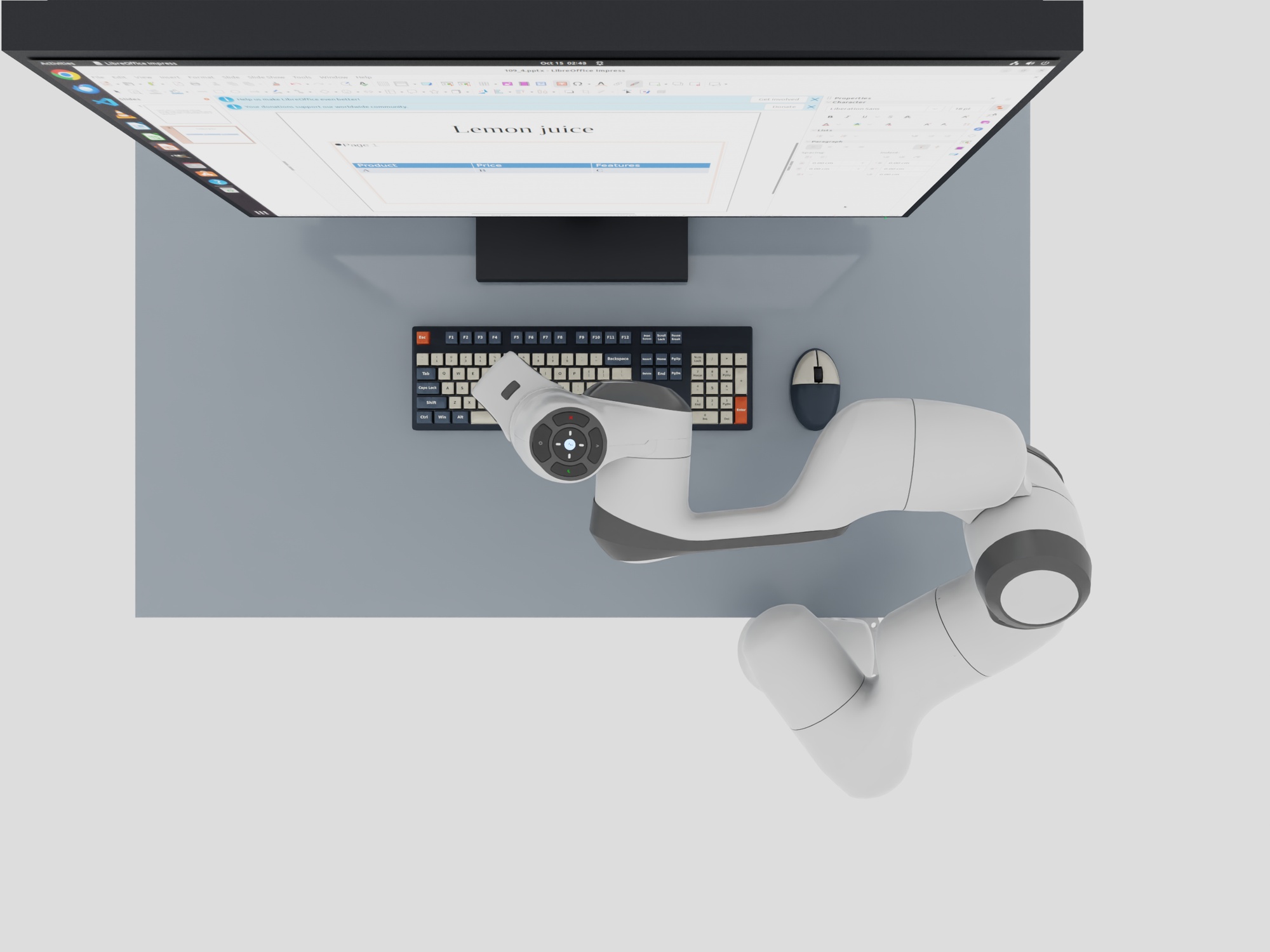}\\
\centering\small (b) Top-down view
\end{minipage}
\caption{Robot-camera observations connect reading the display with locating
the physical controls. The cameras show the screen and controls from oblique and top-down viewpoints, with occasional robot occlusion.}
\label{fig:observations}
\end{figure}

The model receives a front view of the workstation, a top-down view of the
manipulation area, fixed camera calibration, and its own joint, tool-pose,
and contact-force measurements. These observations combine the digital state
visible on the monitor with the physical state needed to act on it.
Figure~\ref{fig:observations} shows the paired camera views.

\begin{table}[t]
\centering\footnotesize
\caption{Control protocol of one decision round. The probe and the dexterous hands share the protocol; only the proprioception fields and the \texttt{robot\_step} arguments differ.}
\label{tab:protocol}
\setlength{\tabcolsep}{4pt}
\begin{tabularx}{\columnwidth}{@{}lX@{}}\toprule
\multicolumn{2}{@{}l}{\textit{Observation}}\\
Task & natural-language instruction, identical to OSWorld\\
Images & current Eye camera (rendered at 3840$\times$2560) and Top camera (2560$\times$1920)\\
Calibration & intrinsics and world-to-camera transform of each camera\\
Proprioception & probe: tip position and orientation, 7 arm joint angles, tip contact force; hands: per arm, wrist and index-finger poses, arm and hand joint angles, commanded hand targets, joint limits, per-link contact forces\\
History & last 3 rounds: proprioception, Eye image and the model's answer\\\midrule
\multicolumn{2}{@{}l}{\textit{Action (one reply)}}\\
Format & JSON with a short \texttt{reason} and 1--8 actions, validated before any motion\\
\texttt{robot\_step} & relative Cartesian target $[d_x,d_y,d_z,r_x,r_y,r_z]$ from the measured pose, lasting 0.05--1.0\,s; hands add absolute targets for the 6 active joints of each hand and may move one or both arms in the same step\\
\texttt{wait} & hold the current targets for 0.05--1.0\,s\\
\texttt{done} & end the episode (last action only)\\
Limits & per action and arm: translation $\le$0.08\,m, rotation $\le$0.35\,rad; per reply: $\le$2.0\,s and $\le$0.24\,m of total translation\\
Budget & up to 100 replies per episode; an invalid reply executes nothing, is returned as feedback, and counts toward the budget\\\bottomrule
\end{tabularx}
\end{table}

Each decision round follows the protocol in Table~\ref{tab:protocol}. The model issues bounded motion commands through a validated JSON interface. A generic inverse-kinematics and joint-drive controller tracks these commands; the model determines contact locations and release timing. Key coordinates, device poses, and direct GUI actions are not provided to the embodied policy.

This design makes the input operation itself part of the task. Pressing a target key
requires spatial alignment and sufficient travel, while a click requires mouse
positioning and button actuation. The same physical action may also displace a
device, unintentionally contact a neighboring key, or obscure the display, changing what the
model must do next.

\benchmark{} evaluates the complete loop from camera observation to application outcome. Task success requires the model to coordinate perception, planning, physical input, and feedback-based recovery under the specified interface and budget; the recorded interactions expose where this end-to-end execution breaks.

\subsection{Physical--digital loop}
Keyboard press and release edges are detected from key travel; mouse movement,
buttons, and wheel motion are detected from peripheral state. Each key is a
sprung prismatic joint that registers a press at 2\,mm of travel and a release
below 1\,mm, with an 8\,ms debounce. The mouse reports motion through an
optical-sensor model (800 DPI) only while it rests within 3\,mm of the desk and
tilts less than 25\textdegree; its buttons actuate at 0.55\,mm and its wheel has
24 detents per turn. A key therefore registers only when contact drives it
past the actuation point, whichever part of the robot makes the contact.

The bridge
sends the generated inputs to the VM. Physics pauses during model inference. To preserve contact durations independently of inference latency, the bridge buffers keyboard and button-held mouse gestures until full release, then replays them using their physical timestamp offsets. Free mouse motion and scrolling are dispatched directly. The policy thus receives gesture-level feedback: press, move as needed, and release, then observe the updated application state together with the measured robot state. Saved images, action sequences, input events, and task outcomes support
analysis of the entire chain from intention to effect.

\section{Paired Evaluation}
\label{sec:protocol}
\subsection{Shared goals and physical control}
We evaluate five models on the same 40 OSWorld tasks, with four tasks in each of ten domains: Chrome, GIMP, LibreOffice Calc, Impress, Writer, multi-application workflows, OS, Thunderbird, VLC, and VS Code. Tasks are selected through stratified sampling, execution-based validation, and manual review to filter out tasks with infrastructure issues or unsolvable steps---verifying that every retained task is solvable in our setting---while maintaining balanced domain coverage and a mix of document editing, application settings, file operations, and cross-application workflows. Each task uses its OSWorld initialization and official evaluator. The selection also spans the input skills a desktop route needs: typing, single keys, key chords, mouse clicks, drags, and scrolling.

The OSWorld condition uses desktop screenshots and PyAutoGUI. The two physical conditions use workstation cameras and either the single-arm rigid probe or the dual-arm hands to operate the keyboard and mouse. Both conditions use the same task definitions and official outcome evaluators, forming one-to-one paired model--task results. Each condition forms a complete observation--action loop: direct desktop perception and software control in the OSWorld condition, and camera perception and contact control in the physical condition.

Both conditions allow up to 100 model decisions. This budget fixes the number of model--environment interaction rounds, not physical motion time or device events; the decisions spent on positioning, contact, and recovery are part of the evaluated embodied task. With API-served models, a 100-step physical episode takes 2--5 hours of wall-clock time, most of it model inference. With full parallelism, simulating the 400 physical episodes reported here occupied 16 RTX 5090 GPUs for about two days; open-weight model inference ran on separate services. Model requests use the maximum reasoning setting and the officially recommended sampling parameters.

\subsection{Outcome measures}
For model $m$ and task $t$, let $s^N_{mt}$ and $s^P_{mt}$ denote the OSWorld and physical scores. ROSWorld reports three quantities per model: the OSWorld (native) score, the physical score, and retention. The first two count successful tasks out of 40 in each condition. Retention is the fraction of tasks successfully completed in OSWorld that are also completed physically:
\begin{equation}
 R_m=\frac{\sum_t\mathbf{1}[s^N_{mt}=1\land s^P_{mt}=1]}{\sum_t\mathbf{1}[s^N_{mt}=1]}.
\end{equation}

\section{What Survives Physical Execution: Results and Analysis}
\label{sec:evaluation}
\begin{table}[t]
\centering\small
\caption{Successful tasks out of 40 (\% in parentheses) in OSWorld and with the single-arm probe and dual-arm hands. Retained: tasks completed successfully in both OSWorld and the physical condition (probe / hands).}
\label{tab:results}
\setlength{\tabcolsep}{3pt}
\resizebox{\columnwidth}{!}{%
\begin{tabular}{lrrrc}\toprule
Model & OSWorld & Probe & Hands & Retained \\\midrule
GPT-6 Astra & 39 (97.5) & 23 (57.5) & 28 (70.0) & 22 / 28 \\
GPT-6.1 Sol & 39 (97.5) & 15 (37.5) & 16 (40.0) & 15 / 16 \\
Claude Opus 5.5 & 39 (97.5) & 11 (27.5) & 3 (7.5) & 11 / 3 \\
Kimi K3 & 31 (77.5) & 0 (0.0) & 0 (0.0) & 0 / 0 \\
GLM-5.3-Flash & 37 (92.5) & 0 (0.0) & 0 (0.0) & 0 / 0 \\\midrule
Total & 185 & 49 & 47 & 48 / 47 \\
\bottomrule\end{tabular}%
}
\end{table}

\subsection{Overall performance}
\begin{figure}[t]
\centering
\includegraphics[width=\linewidth]{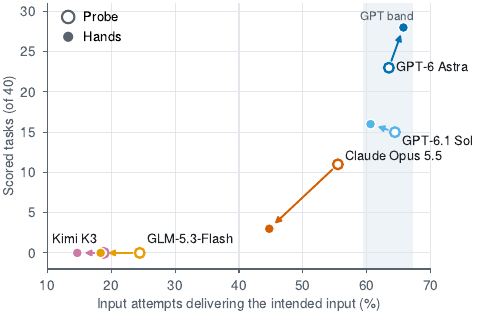}
\caption{Physical success against the share of input attempts that deliver the intended input (as in Figure~\ref{fig:modes}a). Arrows go from the probe (open) to the hands (filled); the shaded band spans the GPT models.}
\label{fig:scatter}
\end{figure}

The five models complete 31--39 of the 40 tasks in OSWorld, whereas physical completion drops sharply and diverges across models (Table~\ref{tab:results}). With the rigid probe, GPT-6 Astra completes 23 tasks, GPT-6.1 Sol 15, and Claude Opus 5.5 11; with the dual-arm hands, they complete 28, 16, and 3, respectively, while Kimi K3 and GLM-5.3-Flash complete none in either embodiment. Physical success is almost strictly a subset of desktop success: 48 of the 49 probe completions and all 47 hand completions also succeed in OSWorld (retention rates of 25.9\% and 25.4\% across the 185 OSWorld completions).

\textbf{Effects of embodiment.} Moving from the probe to the hands lifts GPT-6 Astra from 23 to 28 tasks (8 solved only with the hands, 3 only with the probe), leaves GPT-6.1 Sol nearly unchanged (15 vs.\ 16), and drops Claude Opus 5.5 from 11 to 3. The benefit of added dexterity thus varies across models and comes with additional unintended contacts (Section~\ref{sec:breaks}). Figure~\ref{fig:scatter} relates task success to the share of attempts that deliver the intended device event, with arrows showing the shift between embodiments.

\subsection{Performance across domains}
\begin{figure}[t]
\centering
\includegraphics[width=\linewidth]{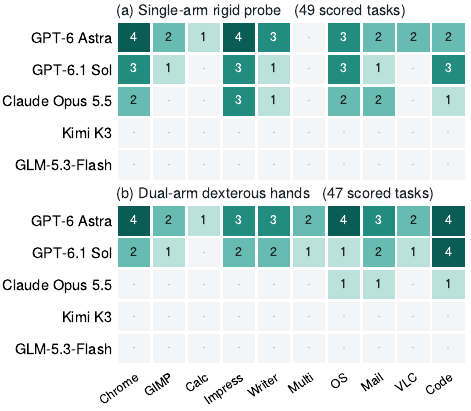}
\caption{Successful model--task pairs by application domain.}
\label{fig:domains}
\end{figure}

Physical completions cover nine of the ten domains with the probe and all ten with the hands (Figure~\ref{fig:domains}). Of the 40 tasks, at least one model completes 25 with the probe, 29 with the hands, and 32 in at least one embodiment. Counting successful model--task pairs, probe completions concentrate in Impress (10), Chrome (9), and OS (8), with none in multi-application workflows. With the hands, GPT-6 Astra succeeds in every domain, including two multi-application tasks; VS Code has the most successful pairs across models (9, compared with 6 for the probe).

\subsection{Route and input modality adaptation}
\label{sec:gui_intent}
\begin{figure}[t]
\centering
\includegraphics[width=\linewidth]{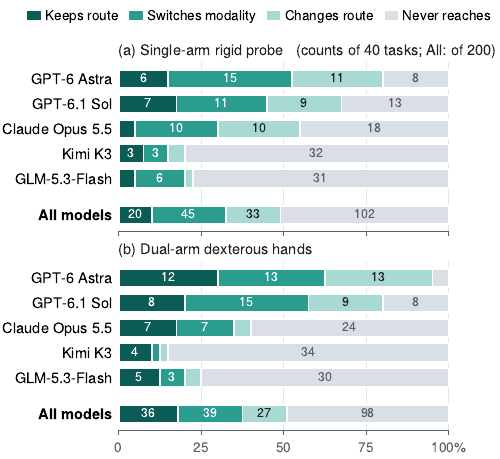}
\caption{Does the OSWorld route survive physical execution? Each model's OSWorld trajectory is aligned with its physical trajectory on the same 40 tasks. \emph{Keeps route}: the operation is reproduced with the same input modality; \emph{switches modality}: the same operation through another input; \emph{changes route}: the physical run gets past the operation by other means; \emph{never reaches}: the run ends before it.}
\label{fig:route_labels}
\end{figure}

We align each model's OSWorld trajectory with its physical trajectory on the same task at the level of semantic operations (e.g., open the Format menu, select the picture), and label every OSWorld operation by what the physical run does with it (Figure~\ref{fig:route_labels}). Two annotators, Qwen3.8-27B and GPT-6.1 Sol, label all 400 pairs and agree on 77\% of task labels ($\kappa=0.66$); disagreements are adjudicated, and hand checks agree on 29 of 32 pairs. The label counts change by at most five pairs when either annotator is used alone.

\textbf{Route preserved; input modality changes.} Among runs that reach the OSWorld route, 65 of 98 with the probe and 75 of 102 with the hands preserve it, either with the original input modality or with a substitute. The most common outcome is the same operation through a different input modality, and the dominant substitution replaces a mouse click with keys (227 operations with the probe, 172 with the hands). None of the 400 pairs reproduces every OSWorld operation unchanged.

\begin{figure}[t]
\centering
\includegraphics[width=\linewidth]{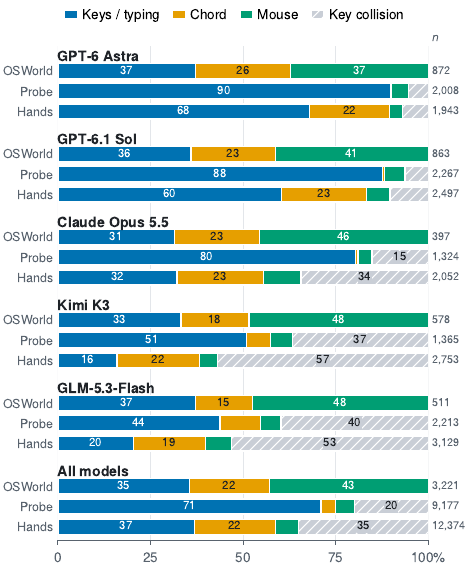}
\caption{Input mix on the desktop and through the robot. OSWorld bars count PyAutoGUI calls (terminal scripts folded into keys/typing); probe and hands bars count input events delivered by the physical devices. \emph{Chord}: a modifier held together with another key; \emph{key collision}: any other key overlap, e.g.\ neighbouring keys struck together. $n$: calls or events.}
\label{fig:modality}
\end{figure}

\begin{figure*}[t]
\centering
\includegraphics[width=\textwidth]{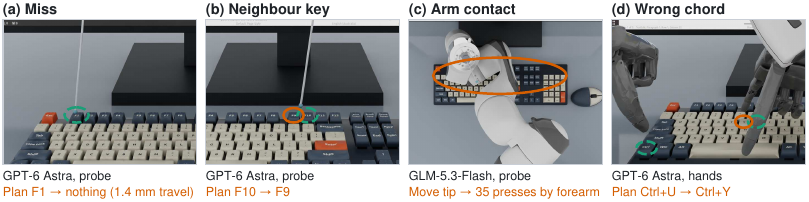}
\caption{Operation-level failures in the robot's own camera views. Dashed green marks the key the model planned; solid red marks what was pressed. (a) The probe rests on F1 but the key travels 1.4\,mm, short of the 2\,mm actuation point. (b) The probe lands between F10 and F9 and presses F9. (c) GLM-5.3-Flash moves its probe tip while the forearm lies on the keyboard and presses keys 35 times; its reply in the same step reports a ``breakthrough'' in arm tracking. (d) For Ctrl+U the left hand holds Ctrl, but the right hand strikes Y instead of U.}
\label{fig:cases}
\end{figure*}

\begin{figure*}[t]
\centering
\includegraphics[width=\textwidth]{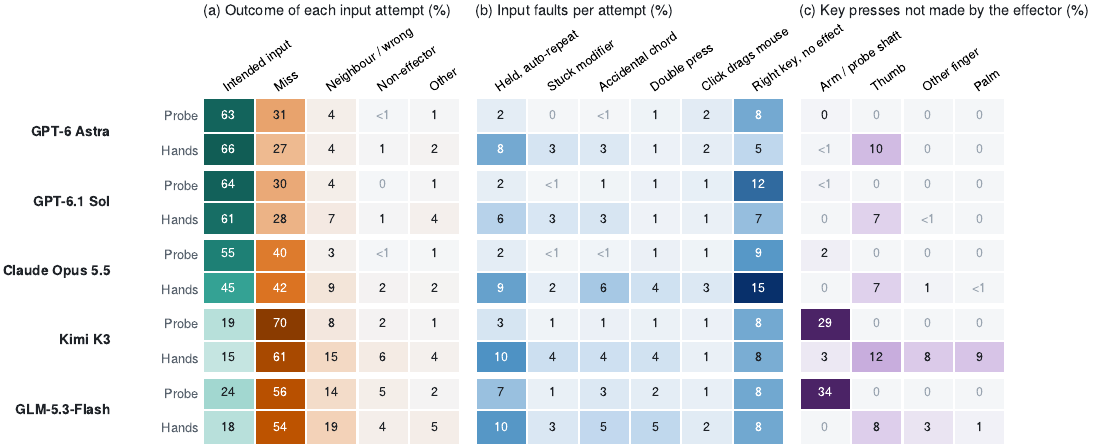}
\caption{Failure modes at the operation level. Rows are model and embodiment; an attempt is an executed step that intends an input. (a) Outcome of each attempt: the intended input arrives, nothing is pressed (miss), a neighbouring or wrong target is pressed, a body part other than the effector presses something, or other input arrives. (b) Input faults per attempt. (c) Share of key presses made by a robot part other than the effector, attributed from the simulator's per-link contact forces; for the probe, presses with the tip more than 3\,cm away or out of contact are assigned to the arm or probe shaft.}
\label{fig:modes}
\end{figure*}

\textbf{Hands restore chords, not the mouse.} Device logs provide a complementary view of the inputs actually delivered (Figure~\ref{fig:modality}). In OSWorld, 43\% of the models' calls are mouse actions and 22\% are chords. Through the robot, mouse input falls to 5.2\% of delivered events with the probe and 6.4\% with the hands, and only 58 of 292 and 68 of 295 reached mouse operations stay on the mouse. Chords split the embodiments: with the probe, only 4 of 80 reached chord operations are still attempted as chords and none arrives as one; with the hands, 88 of 120 reached chord operations are attempted as chords and 81 are delivered. Chords account for 22\% of delivered input events, and 36 tasks retain the original route and input modality, compared with 20 for the probe.

\subsection{Where physical execution breaks}
\label{sec:breaks}
\begin{table}[t]
\centering\footnotesize
\caption{Task-level causes of failure (probe / hands). Out of steps: the 100-step budget is exhausted; gives up: the model ends the run conceding failure or infeasibility; false done: the model declares completion but the evaluator scores zero; other: reasoning length exhausted or a safety refusal.}
\label{tab:task_failures}
\setlength{\tabcolsep}{2.5pt}
\resizebox{\columnwidth}{!}{%
\begin{tabular}{lccccc}\toprule
Model & Failed & Out of steps & Gives up & False done & Other \\\midrule
GPT-6 Astra & 17 / 12 & 16 / 9 & 0 / 0 & 1 / 3 & 0 / 0 \\
GPT-6.1 Sol & 25 / 24 & 23 / 22 & 0 / 0 & 2 / 2 & 0 / 0 \\
Claude Opus 5.5 & 29 / 37 & 23 / 34 & 0 / 0 & 1 / 1 & 5 / 2 \\
Kimi K3 & 40 / 40 & 29 / 30 & 11 / 8 & 0 / 2 & 0 / 0 \\
GLM-5.3-Flash & 40 / 40 & 27 / 26 & 4 / 5 & 1 / 2 & 8 / 7 \\
\midrule
Total & 151 / 153 & 118 / 121 & 15 / 13 & 5 / 10 & 13 / 9 \\
\bottomrule\end{tabular}%
}
\end{table}

We classify every executed step that attempts a key or mouse action (11{,}370 with the probe, 11{,}518 with the hands) by what the devices record, and attribute each key press to the robot part in contact using the simulator's per-link contact forces (Figure~\ref{fig:modes}). Blind checks agree with the labels on 37 of 40 held-out attempts (92.5\%).

\textbf{Missed actuations.} A miss, an intended press that produces no device event, is the most common failure for every model: 27--32\% of GPT attempts and 40--70\% for the others. With the probe, most misses are hovers in which the tip touches nothing (4{,}509 of 5{,}365); with the hands, most land in a key gap or on the case without moving a key (3{,}063 of 5{,}025). Misses expose failures in effector-to-target alignment (Figure~\ref{fig:cases}a).

\textbf{Wrong targets and actuation faults.} Neighbouring or wrong keys stay at 4--7\% of GPT attempts in both embodiments but rise from 3\% to 9\% for Claude Opus 5.5 and to 15--19\% for Kimi K3 and GLM-5.3-Flash with the hands (Figure~\ref{fig:cases}b). The hands also raise actuation faults: keys held long enough to auto-repeat reach 6--10\% of attempts (2--7\% with the probe), and modifiers left down and accidental chords become more frequent.

\textbf{Whole-body awareness.} With the probe, 29--34\% of the keys Kimi K3 and GLM-5.3-Flash press are struck by the arm or the probe shaft, not the tip, against at most 2\% for the other models. With the hands, Kimi K3 presses 18\% of keys with fingers other than the pointing one or with the palm. These are not positioning errors at the target but a missing model of the whole embodiment: in Figure~\ref{fig:cases}c the model reports a breakthrough in arm tracking while its forearm presses keys 35 times, and in Figure~\ref{fig:cases}d the right hand turns Ctrl+U into Ctrl+Y.

\textbf{Application-state mismatch.} In 5--15\% of attempts, the intended key is delivered but does not produce the expected effect because focus or application state differs from what the model assumed.

\textbf{Task termination.} Of 304 failed episodes, 239 exhaust the 100-step budget (Table~\ref{tab:task_failures}). Of the budget-exhausted runs, 217 complete no new OSWorld route operation during the final 20 decisions, and 166 include at least three consecutive failed attempts at the same target. Such retries fill 7--13\% of the final attempts for the GPT models and 53--77\% for Kimi K3 and GLM-5.3-Flash. Successful runs use a median of 43 decisions. Only Kimi K3 and GLM-5.3-Flash give up (28 episodes). False completions, where the model declares success and the evaluator disagrees, are rare (15); in 6 of 10 sampled cases the change was made on screen but never saved or exported.

\subsection{Decision budget}
\label{sec:budget}
\begin{figure}[t]
\centering
\includegraphics[width=\linewidth]{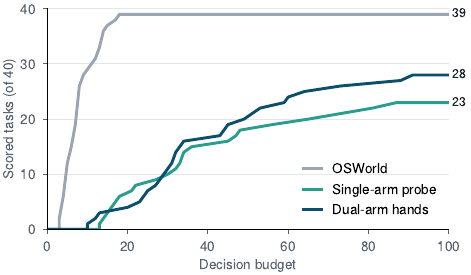}
\caption{GPT-6 Astra: successful tasks out of 40 by decision budget in OSWorld and with each embodiment. Curves are computed from recorded runs.}
\label{fig:budget}
\end{figure}

Figure~\ref{fig:budget} shows the decision-budget curve for GPT-6 Astra, the strongest physical model. All 39 OSWorld completions finish within 18 decisions (median 8), whereas physical completions require a median of 33 with either embodiment. Thus, the 100-decision limit already exceeds five times its longest OSWorld run. By 50 decisions, the probe and hands solve 18 and 20 tasks; by 80, they reach 21 and 26, gaining only two more each by 100 (23 and 28). Physical execution demands far more rounds, yet gains flatten well before the limit.

\subsection{Implications for model training}
Each failure mode points to data that desktop trajectories do not contain.

\textbf{Effector-to-target geometry.} Misses are the largest failure for every model (27--70\% of attempts). Training data should pair camera views and effector poses with whether, and which, key actuated, including near misses, so that a model learns where its effector is relative to a target.

\textbf{Whole-body awareness.} Up to a third of the weaker models' key presses come from the arm, the probe shaft, or a finger other than the one pointing. Data should expose the contacts of the entire body, not only the effector, and favour plans that keep the rest of the body clear of the device.

\textbf{Embodiment-conditioned routes and verification.} Models keep their desktop routes but must work around actions the effector cannot deliver, and even the strongest models stall on repeated, unverified retries. Learning which input modalities an embodiment can produce, and confirming each effect on screen before the next step, would turn the same desktop competence into fewer and surer physical steps.

\section{Conclusion}
\benchmark{} evaluates physical computer use through shared desktop goals, official evaluators, and robot contact with a keyboard and mouse. Models with strong desktop scores show sharply different physical completion, and the choice of embodiment further separates them. Paired trajectories trace the gap from model plans through contact and input events to application outcomes, establishing \benchmark{} as a testbed for turning digital task competence into embodied action.

\FloatBarrier

\bibliographystyle{ieeenat_fullname}
\bibliography{references}

\end{document}